\documentclass[letterpaper]{article} % For LaTeX2e
\usepackage[preprint]{colm2026_conference} % preprint, final, submission

\usepackage{wrapfig}
\usepackage[utf8]{inputenc} % allow utf-8 input
\usepackage[T1]{fontenc}    % use 8-bit T1 fonts
\usepackage{hyperref}       % hyperlinks
\usepackage{url}            % simple URL typesetting
\usepackage{booktabs}       % professional-quality tables
\usepackage{amsfonts}       % blackboard math symbols
\usepackage{nicefrac}       % compact symbols for 1/2, etc.
\usepackage{microtype}      % microtypography
\usepackage{xcolor}         % colors
\usepackage{graphicx}
\usepackage{enumitem}
\usepackage{subcaption}
\usepackage{multirow}
\usepackage{tcolorbox}
\tcbuselibrary{listings, breakable}
\usepackage{amsmath}
\usepackage{amssymb}

\usepackage{lineno}

\definecolor{darkblue}{rgb}{0, 0, 0.5}
\definecolor{seedgray}{gray}{0}

\hypersetup{colorlinks=true, citecolor=darkblue, linkcolor=darkblue, urlcolor=darkblue}

\newtcblisting{promptbox}[1][]{
  colback=gray!5,
  colframe=gray!50,
  listing only,
  unbreakable,
  listing options={
    basicstyle=\ttfamily\small,
    breaklines=true,
    breakautoindent=false,
    breakindent=0pt
  },
  title=#1,
  fonttitle=\bfseries\sffamily
}
\newtcolorbox{wipbox}[1][]{
    colback=gray!5,
    colframe=gray!50,
    fonttitle=\bfseries,
    title=WIP: #1,
    sharp corners,
    boxrule=0.5pt
}
\tcbuselibrary{skins}

\lstdefinestyle{prompt}{
  basicstyle=\scriptsize\ttfamily,
  breaklines=true,
  breakatwhitespace=false,
  numbers=left,
  numberstyle=\tiny\color{gray},
  numbersep=8pt,
  keepspaces=true,
  columns=fullflexible,
  frame=none,
  xleftmargin=18pt,
  escapeinside={(*@}{@*)},
  breakindent=0pt,
}

\title{Agents Explore but Agents Ignore: LLMs Lack Environmental Curiosity}

\author{Leon Engl{\"a}nder \\
Cohere \\
\texttt{leon@cohere.com} \\
\And
Sophia Althammer \\
Cohere \\
\And
Ahmet {\"U}st{\"u}n \\
Cohere \\
\And
Matthias Gall{\'e} \\
Poolside \\
\And
Tom Sherborne \\ 
Cohere
}

\begin{document}

\ifcolmsubmission
\linenumbers
\fi

\maketitle

\begin{abstract}
LLM-based agents are assumed to integrate environmental observations into their reasoning: discovering highly relevant but unexpected information should naturally lead to a model exploiting its own discoveries. We show that this assumption is false for current LLM-based agents, which struggle to reflect or react to unexpected information.
Across three benchmarks (Terminal-Bench, SWE-Bench, AppWorld), we inject complete task solutions into the agent environments to deliberately expose a task's solution to a model. While agents discover these solutions on Terminal-Bench in 79–81\% of runs, they interact, or exploit, them in only 37-50\% of cases. This gap is starkest in AppWorld: agents see documentation stating that a command ``returns the complete solution to this task'' in over 90\% of attempts but exploit this in fewer than 7\% of trials.
We show that agents lack what we call \emph{environmental curiosity}: the capability to recognize and investigate unexpected but relevant observations in response to environmental stimuli. We identify three main factors influencing environmental curiosity: available tools in the agent scaffold, test-time compute, and training data distribution. Our findings identify configurations that maximize curiosity also achieve the best performance on the unmodified benchmarks. Yet even jointly optimized agents still ignore discovered solutions in the majority of trials: current agents use the environment to fetch expected information, but not to revise their strategy or maximally exploit useful stimuli.

\end{abstract}

\section{Introduction}
Contemporary LLM-based agents have made rapid progress on benchmarks simulating complex real-world tasks.
On SWE-Bench Verified \citep{swebenchverified}, resolution rates climbed from 33.2\% to 80+\% through improved models and agent scaffolds like SWE-Agent \citep{sweagent} and OpenHands \citep{openhands}.

Agents begin each task without knowledge of the codebase they act in. To succeed, they must explore the environment to discover relevant information and integrate findings into their reasoning and future steps. During exploration, the agent may encounter unexpected but highly relevant information, such as code that already solves parts of the task. In those situations, an agent would benefit from environmental curiosity, which we define as the capability to recognize and investigate such observations in response to environmental stimuli. We show that current agents lack this useful exploration behavior: while agents regularly discover relevant but unexpected information, they fail to investigate and apply this information to solve the problem at hand, but rather ignore the information.

To evaluate environmental curiosity, we propose \emph{solution injection}: we place the respective solution directly inside the environment, e.g., as a script, and measure \textit{(i)} whether agents discover the solution and \textit{(ii)} whether they interact with it, e.g., read the added solution file. Existing benchmarks primarily measure task success (often based on environment state), and thus cannot distinguish agents that adapt their behavior based on observations from those that execute fixed patterns learned during training. Through solution injection, we are able to quantify this distinction.

\begin{figure}
    \centering
    \includegraphics[width=\linewidth]{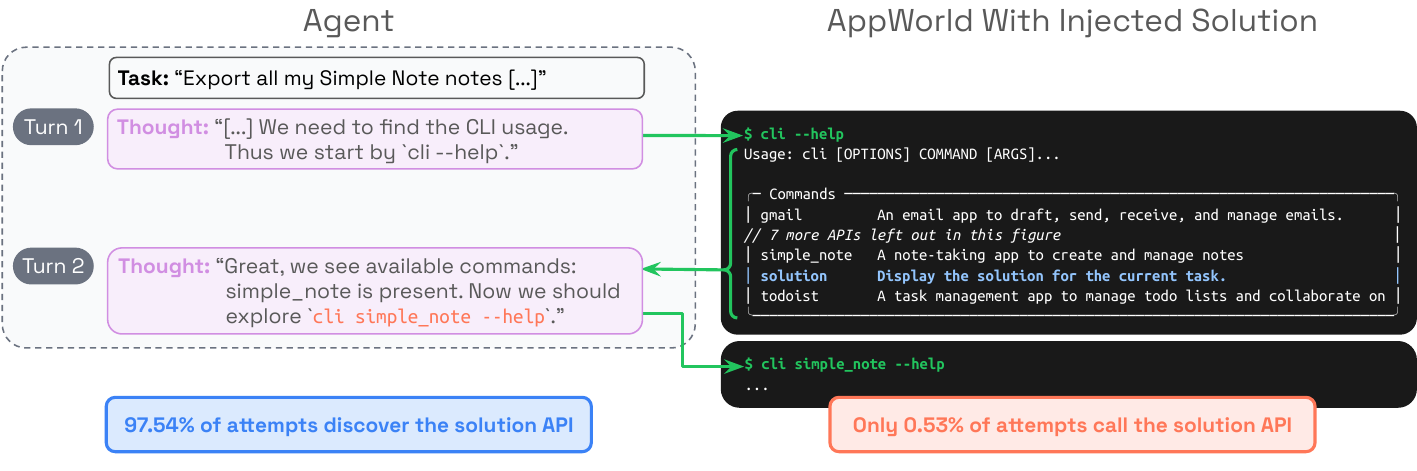}
    % \vspace{-1.75em}
    \caption{Agents discover but ignore injected solutions. This figure shows a trajectory from AppWorld with injected solution using \texttt{gpt-oss-120b} as the LLM. The agent executes \texttt{cli {-}{-}help} and observes documentation explicitly stating that calling the solution API would display the solution for the current task. The agent proceeds without calling it. Current LLM-based agents lack environmental curiosity: while \texttt{gpt-oss-120b} discovers the documentation in 97.54\% of runs, it calls the solution API in only 0.53\% of cases.}
    \label{fig:appworld-example}
    % \vspace{-1.75em}
\end{figure}

We apply solution injection to three agentic benchmarks: Terminal-Bench \citep{terminal-bench}, SWE-Bench Verified \citep{swebenchverified}, and AppWorld \citep{appworld} that span code and non-code domains, including terminal tasks, software engineering tasks, and everyday digital tasks performed via API calls. Our experiments show that across all three benchmarks, and two distinct agent scaffolds (SWE-agent, \citealp{sweagent}; Terminus, \citealp{terminal-bench}), agents frequently discover injected solutions but rarely interact with them. This discovery-interaction gap is starkest in AppWorld: in 97.54\% of attempts, \texttt{gpt-oss-120b} sees documentation explicitly stating that a command ``returns the complete solution to this task'', yet calls this tool only 0.53\% of the time, as illustrated in Figure~\ref{fig:appworld-example}.

We find three critical factors that strongly influence environmental curiosity at inference time: \textbf{tool availability}, \textbf{reasoning budget}, and \textbf{exploration-oriented prompting}. Adding tools beyond a basic \texttt{bash} shell strongly reduces interaction rates, as agents default to learned tool-specific patterns rather than examining their environment. Yet even with all those factors optimized, agents still ignore discovered solutions in the majority of trials, indicating that the deficit is not solely in inference-time configuration. We further investigate the role of training distribution and find that supervised fine-tuning on narrow, in-distribution data reduces environmental curiosity and the diversity of explored solution paths, degrading the benefit of increased performance from additional trials, seen in lower pass@$k$ for higher $k$.

Crucially, the prompts that most improve interaction rates also achieve the best performance on the original, unmodified benchmarks, and narrow in-distribution fine-tuning degrades pass@$k$ scaling on the original benchmarks as well. Optimizing for environmental curiosity consistently improves performance on the original benchmarks.

Our \textbf{main contributions} are:
\begin{itemize}[leftmargin=1.5em, itemsep=0.1em, topsep=0pt, partopsep=0pt]
    \item We define \textbf{Environmental Curiosity} and benchmark the (lack of) capacity of modern LLMs to leverage unexpected relevant information. This gap between discovery and interaction is consistent across three benchmarks, multiple LLMs, and agent scaffolds.
    \item We introduce \textbf{Solution Injection} as a method for adapting agent benchmarks to evaluate environmental curiosity. We propose two new metrics, discovery@$k$ and interaction@$k$, to separately measure whether agents discover and act on relevant information.
    \item We investigate test-time factors influencing environmental curiosity. The most consistent effect comes from tool availability: restricting the agent scaffold to bash-only roughly doubles interaction rates. Increased reasoning budget and prompts instructing to explore also improve environmental curiosity. Yet even with all factors optimized, agents still ignore discovered solutions in the majority of runs.
    \item We find that fine-tuning on narrow in-distribution data reduces environmental curiosity and the diversity of explored solution paths, leading to worse pass@k scaling even on the original benchmarks.
\end{itemize}

\section{Method}
To evaluate environmental curiosity, i.e., the tendency to investigate relevant observations in response to environmental stimuli, we need to separately measure an agent's ability to discover relevant information and whether it interacts with what it discovers. To achieve this, we propose injecting the task's gold solution directly into the environment.

\subsection{Solution Injection}
The central idea is to add the task solution directly to the environment. This is conceptually applicable to any existing agent benchmark where a gold solution exists. The injected solution must be \textit{(i)} complete so that following it guarantees task success and \textit{(ii)} discoverable through agent actions.

This offers unexpected but highly relevant information. It is relevant because it contains the complete task solution, and unexpected because it lies outside the agent's typical workflow. This allows us to measure whether the agent is environmentally curious enough to investigate it. The injected solutions are deliberately obvious. Agents that ignore information explicitly labeled as the solution are unlikely to integrate the subtler information present in real environments. We apply solution injection to Terminal-Bench, SWE-Bench, and AppWorld, injecting solutions as executable files or as documented API endpoints, depending on the benchmark\footnote{For example, in Terminal-Bench and SWE-Bench, we add the solution as \texttt{solution.sh} to the agent's working directory. In AppWorld, we add a \texttt{solution} API endpoint documented in the CLI help output.}.
To ensure our findings are not artifacts of a specific file name or format, we also experiment with different injection variants (Appendix~\ref{appendix:investigating_different_solution_file_names}, \ref{appendix:more_difficulty_levels_of_solution_injection}).

\subsection{Metrics}
To measure performance and how often the agent discovers and exploits solution injection, we measure three metrics across $n$ attempts per task. We use the pass@k definition from \citet{chen2021evaluating_llms_trained_on_code} and introduce two new metrics for discovery and interaction using the same unbiased estimator to compute probabilities across $n$ attempts:

\textbf{pass@$k$}: The probability that at least one of $k$ attempts successfully completes the task. With $c_{\text{pass}}$ being the number of attempts that pass the task:
\begin{equation}
    \text{pass@}k := \underset{\text{Tasks}}{\mathbb{E}}\left[1 - \dfrac{\binom{n - c_{\text{pass}}}{k}}{\binom{n}{k}}\right] \label{eq:discovery_at_k}
\end{equation}

\textbf{discovery@$k$}: The probability that at least one of $k$ attempts executes a command that surfaces the injected solution in the agent's context. This metric serves as a sanity check that the injected solution is indeed discoverable through the normal agent's actions. With $c_{\text{disc}}$ being the number of attempts in which the solution was discovered:
\begin{equation}
    \text{discovery@}k := \underset{\text{Tasks}}{\mathbb{E}}\left[1 - \dfrac{\binom{n - c_{\text{disc}}}{k}}{\binom{n}{k}}\right] \label{eq:discovery_at_k}
\end{equation}

\textbf{interaction@k}: The probability that, across $k$ attempts, the agent interacts with the injected solution at least once, such as reading or executing the solution file or querying the solution API.\footnote{We detect interaction by checking whether any command executed by the agent references the injected solution, e.g. contains ``\texttt{solution.sh}'' or ``\texttt{cli solution}''.}
This metric measures environmental curiosity: a high interaction rate would mean that the agent investigated the unexpected but highly relevant information it discovered. A low interaction rate would mean that it ignored it. 
With $c_{\text{interact}}$ being the number of attempts in which the agent interacted with the solution, we define $\text{interaction@}k$ as:
\begin{equation}
    \text{interaction@}k := \underset{\text{Tasks}}{\mathbb{E}}\left[1 - \dfrac{\binom{n - c_{\text{interact}}}{k}}{\binom{n}{k}}\right]  \label{eq:interaction_at_k}
\end{equation}

\section{Experiments}
We evaluate the environmental curiosity of agents and investigate which factors influence environmental curiosity. We structure our investigation around three hypotheses:
% \begin{itemize}
%     \item[\textbf{(H1)}]~LLM-based agents lack environmental curiosity: they discover relevant information during exploration but systematically fail to act on it.\\
%     \item[\textbf{(H2)}]~Test-time design decisions (e.g., prompt, instruction following) are the primary factors behind environmental curiosity.
%     \item[\textbf{(H3)}]~Narrow domain fine-tuning minimizes environmental curiosity as a trade-off for in-domain performance gain.
% \end{itemize}
\textbf{(H1)} LLM-based agents lack environmental curiosity: they discover relevant information during exploration but systematically fail to act on it. \textbf{(H2)} Test-time design decisions shape environmental curiosity. \textbf{(H3)} Narrow fine-tuning suppresses environmental curiosity.

\paragraph{Benchmarks.} We apply solution injection to three benchmarks: Terminal-Bench v1~\citep{terminal-bench}, SWE-Bench Verified~\citep{swebenchverified}, and AppWorld~\citep{appworld}. Terminal-Bench spans a wide variety of terminal-based tasks, including file manipulation, system administration, and data processing. SWE-Bench Verified evaluates agents on resolving real GitHub issues by editing repository code. AppWorld requires agents to complete everyday digital tasks, such as managing emails, notes, and calendars, by interacting with simulated apps via API calls. For Terminal-Bench and SWE-Bench Verified, we inject the solution as an executable \texttt{solution.sh} in the agent's working directory. For AppWorld, we add a \texttt{solution} API endpoint documented in the ``\texttt{cli {-}{-}help}'', as shown in Figure~\ref{fig:appworld-example}. We also report on AppWorld's \textit{validation} split, rather than \textit{test} split, as we require gold solutions for solution injection, which only exist for the validation split.

\paragraph{Agent implementation.} The agent setup involves three layers: the \emph{harness}, \emph{scaffold}, and \emph{tools}. The \emph{harness} handles the execution environment, i.e., initializes the Docker execution environment and controls evaluating submitted solutions. The \emph{scaffold} is the agent loop (i.e., ReACT loop from \citet{react}), including prompting, tool-call parsing, and history management. The \emph{tools} determine how the agent may interact with its environment. We evaluate two scaffolds: Terminus~1~\citep{terminal-bench}, which is the default Terminal-Bench agent, and SWE-agent~\citep{sweagent}. We adapt these scaffolds to use native function-calling APIs over raw prompting to remove the potential variable of out-of-distribution function calling interfaces (introduced in proprietary scaffolds) to instead rely on a provider's native tool-use interface.

We evaluate two tool suites: \texttt{bash}-only, and \texttt{bash} and \texttt{str\_replace\_editor}. 
The \texttt{str\_replace\_editor} is a structured file-editing tool introduced by Anthropic~\citep{anthropic2024computeruse}\footnote{\url{https://platform.claude.com/docs/en/agents-and-tools/tool-use/text-editor-tool}} that has become the standard editing tool in coding agent scaffolds~\citep{sweagent, openhands}. 
This is one of only three tools in SWE-agent's default configuration, alongside \texttt{bash} and a submit tool.\footnote{For SWE-agent, ``bash-only'' refers to \texttt{bash} plus the submit tool, i.e.\ without \texttt{str\_replace\_editor}.}

\paragraph{Evaluation setup.} We evaluate three LLMs: \texttt{gpt-oss-120b}~\citep{gptoss} with high reasoning, \texttt{GLM-4.7}~\citep{glm45}, and fine-tuned variants of \texttt{command-a-reasoning}~\citep{command_a, cohere2025commandareasoningblog} trained on different task distributions, as described in Section~\ref{sec:sec_3_3_effect_of_training_distribution}. Unless otherwise specified, we use the Terminal-Bench harness and the Terminus agent with \texttt{bash} as the only tool. We evaluate with $n = 10$ attempts per task and report discovery@$k$ and interaction@$k$ (Equations~\ref{eq:discovery_at_k} and \ref{eq:interaction_at_k}) alongside pass@$k$ for all experiments.

\begin{figure}[t]
    \centering
    \includegraphics[width=1\linewidth]{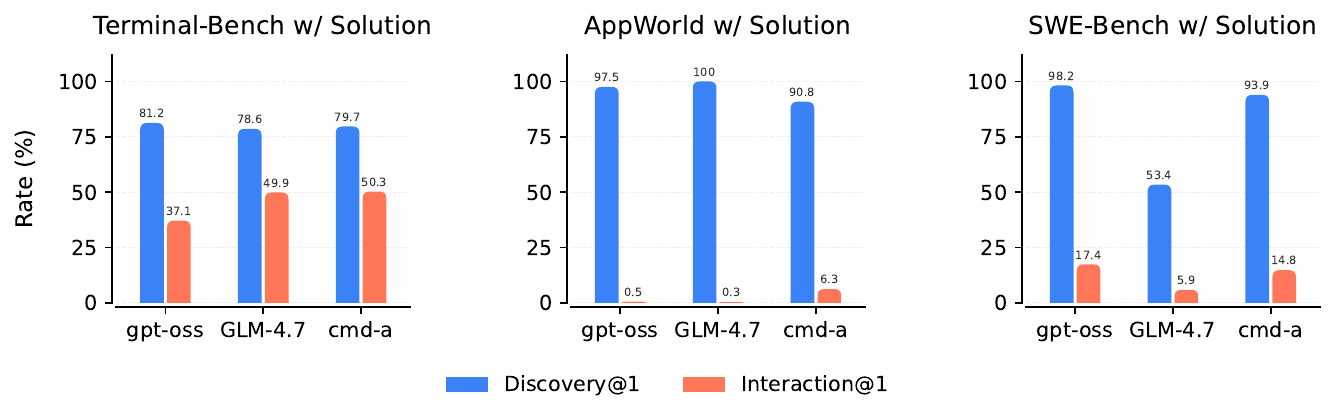}
    % \vspace{-1.75em}
    \caption{discovery@1 versus interaction@1 across benchmarks. We evaluate \texttt{gpt-oss-120b} with high reasoning (gpt-oss), GLM-4.7, and Command A Reasoning fine-tuned for Terminal-Bench (\texttt{cmd-a}). Agents consistently discover solutions but rarely interact with them.}
    \label{fig:discovery-interaction-bar-chart}
\end{figure}

\subsection{Agents lack environmental curiosity}
Figure~\ref{fig:discovery-interaction-bar-chart} shows the discovery and interaction rates across all three benchmarks. Across all models and benchmarks, agents consistently discover the injected solutions but rarely interact with them. On Terminal-Bench, discovery@1 ranges from 78.6--81.2\% while interaction@1 reaches only 37.1--50.3\%. On SWE-Bench, agents discover solutions in 53.4--98.2\% of runs but interact in only 5.9--17.4\%. The gap is starkest on AppWorld: discovery@1 exceeds 90\% for all models, yet interaction@1 never surpasses 6.3\%. The bottleneck for current agents is not discovering relevant information but integrating observations into their reasoning. We show an example trajectory in Figure~\ref{fig:appworld-example}.

\begin{wraptable}{r}{0.4\textwidth}
    \centering
    \vspace{-\intextsep}
    \resizebox{0.4\textwidth}{!}{
\begin{tabular}{@{}lccc@{}}
\toprule
Benchmark & Original & w/ Solution & $\Delta$ \\
\midrule
Terminal-Bench & 44.5 & 55.9 & +11.4 \\
AppWorld & 40.5 & 43.1 & +2.6 \\
SWE-Bench & 45.9 & 46.9 & +1.0 \\
\bottomrule
\end{tabular}
}
    \caption{Task performance of \texttt{gpt-oss-120b} (high reasoning) on original and solution-injected benchmarks. Improvements correlate with the interaction rate in Figure \ref{fig:discovery-interaction-bar-chart}.}
    \label{tab:gpt_oss_pass_rates}
    \vspace{-\intextsep}
\end{wraptable}

Table~\ref{tab:gpt_oss_pass_rates} shows the performance on the original and solution-injected benchmarks for \texttt{gpt-oss-120b}. On the original benchmarks, the model achieves 40-46\% pass@1, which confirms that these are challenging tasks. Injecting solutions improves performance, with the largest gain on Terminal-Bench (+11.4) where interaction is highest. On AppWorld, where agents almost never call the solution API, the gain is minimal (+2.6). This observation holds across all models; full results in Appendix~\ref{appendix:sec:all_models_results}.

\subsection{Test-time factors}
\label{sec:test_time_factors}
We next investigate why agents fail to use what they discover. We begin with examining how test-time design decisions shape environmental curiosity \textbf{(H2)} by investigating three factors: agent scaffolding, test-time compute, and prompting.

\paragraph{Agent scaffolding.}
We compare Terminus and SWE-agent scaffolds to ensure our findings are not artifacts of any specific agent implementation. This comparison additionally investigates how available tools shape environmental curiosity by evaluating two tool configurations: \texttt{bash}-only, and \texttt{bash} with \texttt{str\_replace\_editor}. We evaluate \texttt{gpt-oss-120b} and a fine-tuned variant of \texttt{command-a-reasoning} trained on \texttt{bash}-only trajectories from SWE-smith (\texttt{SWE-Bench-SFT}; described in Section~\ref{sec:sec_3_3_effect_of_training_distribution}). All experiments are conducted on SWE-Bench Verified; for scaffold implementation details, see Appendix~\ref{appendix:agent_implementations}.
Figure~\ref{fig:scaffolding} shows that scaffold choice substantially affects both task performance and environmental curiosity (e.g., \texttt{SWE-Bench-SFT} with \texttt{bash}-only achieves $46.9$ pass@1 with Terminus but $25.2$ with SWE-agent), but both scaffolds consistently exhibit the discovery-interaction gap.

The available tools substantially affect environmental curiosity. Adding \texttt{str\_replace\_editor} improves pass@1 across all configurations but reduces interaction, conditional upon discovery, in every case. The same pass@1 improvement appears on the original, unmodified SWE-Bench Verified, confirming that the performance gains from richer toolsets are not artifacts of solution injection. Notably, \texttt{SWE-Bench-SFT} has never encountered \texttt{str\_replace\_editor} during fine-tuning, yet providing it at test time still relatively reduces interaction given discovery by 13.7\% to 48.3\%, suggesting that the tool-use patterns that reduce exploration originate from pre-training. Among trajectories where agents do interact with the solution, the median interaction step is similar with (step~18) and without (step~15) the editor. Both medians fall within the first 25\% of the trajectory, indicating that the tool does not delay interaction but suppresses it. We conjecture that when a dedicated editing tool is available, agents default to applying it directly rather than first examining their environment; we discuss this further in Section~\ref{sec:discussion}.

\begin{figure}
    \centering
    \includegraphics[width=\linewidth]{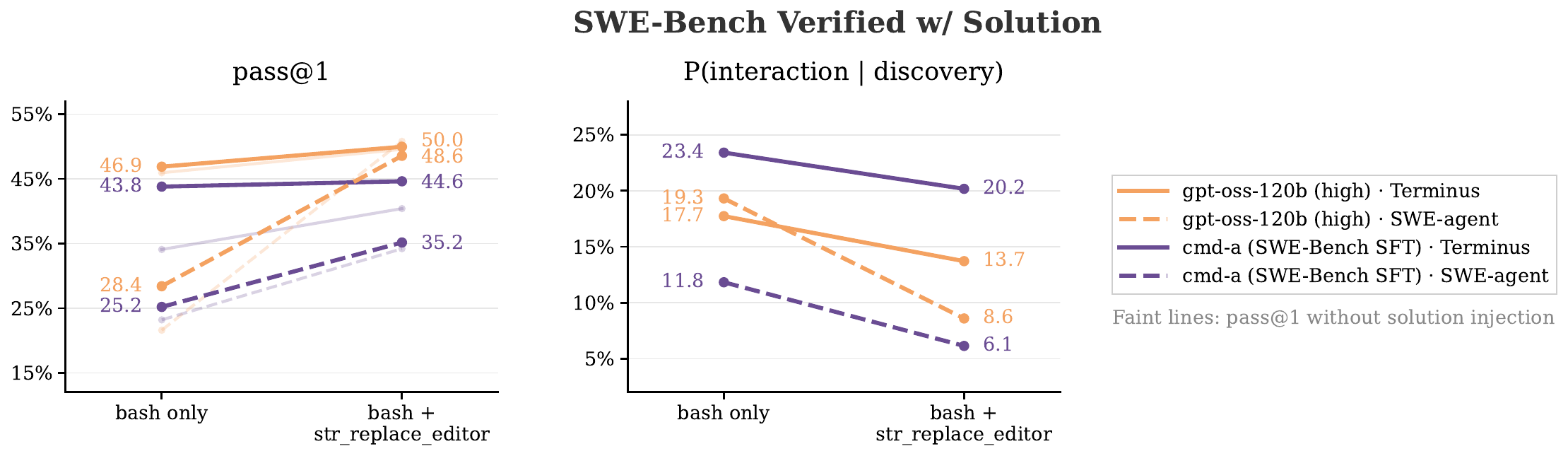}
    \caption{Adding tools improves task performance but reduces environmental curiosity. We evaluate \texttt{gpt-oss-120b} and \texttt{cmd-a} (\texttt{SWE-Bench SFT}) across two scaffolds (Terminus, SWE-agent) on SWE-Bench Verified. Adding \texttt{str\_replace\_editor} increases pass@$1$ for the solution injected as well as the original (faint lines) benchmark, but consistently decreases the probability of interacting with discovered solutions. The discovery@1 exceeds 87.9\% across all configurations, i.e., the conditional metric is not driven by varying discovery rates.}
    \label{fig:scaffolding}
\end{figure}

\paragraph{Effect of reasoning effort.}

\begin{wrapfigure}{r}{0.35\linewidth}
    \centering
    \vspace{-\intextsep}
    \includegraphics[width=\linewidth]{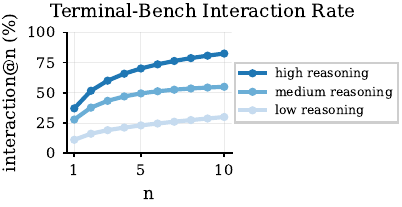}
    % \vspace{-1.5em}
    \caption{\texttt{gpt-oss-120b} with different reasoning budgets on Terminal-Bench.}
    \label{fig:gpt_reasoning_levels_tbench_interaction_rate}
    \vspace{-\intextsep}
\end{wrapfigure}

To additionally measure the effect of reasoning levels, we evaluate \texttt{gpt-oss-120b} with low, medium and high reasoning on all benchmarks. 
We observe that increased reasoning substantially improves environmental curiosity: Terminal-Bench interaction@1 more than triples from 11\% to 37\%, as shown in figure~\ref{fig:gpt_reasoning_levels_tbench_interaction_rate}. 
This is not an artifact of higher discovery rates: The conditional probability of interaction given discovery increases from 17.65\% (low) to 36.68\% (medium) to 45.69\% (high). %, confirming that additional compute helps agents act on information they have already observed, not just observe more.
On SWE-Bench, interaction@1 increases from 0.78\% to 17.42\% when increasing reasoning. On AppWorld, interaction remains near zero across all reasoning levels, indicating that increased compute alone cannot overcome the gap for all task types.

\paragraph{Prompting.}
We test whether explicit instructions improve environmental curiosity. Prompting the agent to explore its environment before acting improves task performance across all three benchmarks by on average +2.57\% on the original benchmarks and +2.96\% on the solution-injected versions, as shown in Appendix~\ref{appendix:prompt_variations_exploration}. On Terminal-Bench, we further evaluate prompts encouraging reflection and adapting to environmental observations (Appendix~\ref{appendix:prompt_variations_on_tbench}). Less directive prompts to encourage curiosity and step-wise reflection yield diminishing returns. Requiring agents to investigate all discovered files before proceeding achieves the highest interaction and pass rates on the solution injected benchmark. Notably, the best-performing prompt on the solution-injected benchmark is also the best-performing prompt on the original Terminal-Bench, suggesting that improved environmental curiosity benefits general task performance.

% NOte from tom: It would be good to briefly mention instruction -following here as a prompt following capability -- just to acknowledge that exploration/curiousity CAN just be an instruction following consequence but we find this is insufficient. 

\paragraph{Summary.}
All three test-time factors improve interaction rates as proposed by Hypothesis \textbf{(H2)}. Tool availability shapes how agents approach their environment: with fewer tools, agents must examine files to understand the environment; with more tools, they default to learned tool-specific patterns and skip investigation or curiosity. Test-time compute and prompt design support agents to identify and act on unexpected but highly relevant information. Yet even with optimal settings, i.e., bash-only, high reasoning, and explicit instructions to investigate, agents still ignore solutions in the majority of trials. This suggests the limitation is not solely a matter of inference-time configuration, but is inherent to how LLMs are trained. Given this finding, we now examine the influence of training data distribution on how agents exhibit curiosity-driven behavior.

\subsection{Effect of training distribution}
\label{sec:sec_3_3_effect_of_training_distribution}
The previous section identified that optimizing test-time factors does not resolve the gap between discovery and interaction, suggesting that the limited environmental curiosity stems from the training phase. To investigate this, we fine-tune \texttt{command-a-reasoning} via rejection sampling~\citep{deepseek_r1} on three task distributions: Terminal-Bench-like tasks from an external vendor (\texttt{T-Bench-SFT}), the AppWorld training split (\texttt{AppWorld-SFT}), and SWE-smith (\texttt{SWE-Bench-SFT}). All models are trained on approximately 20,000 turns; further details are provided in Appendix~\ref{appendix:sft_training_data_details}. We use these models to analyze how training breadth and domain transfer affect environmental curiosity.

\begin{figure}[t]
    \centering
    \includegraphics[width=0.49\textwidth]{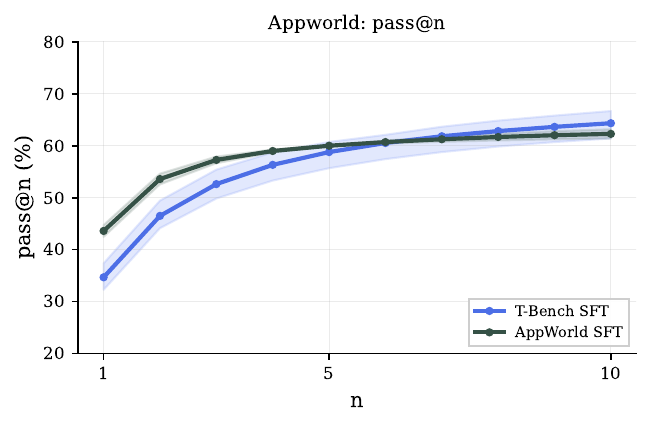}
    \includegraphics[width=0.49\textwidth]{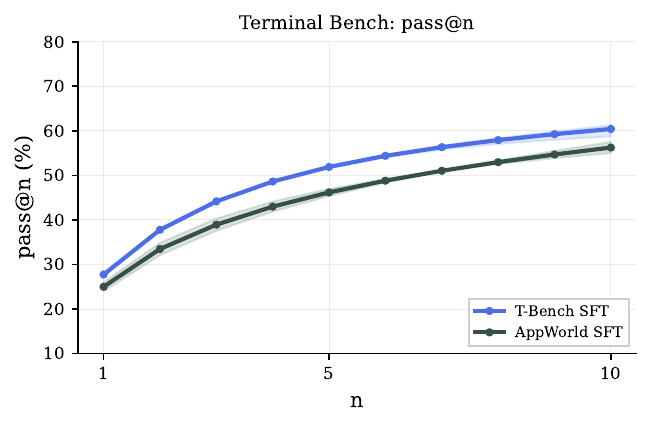}
    \caption{Training distribution affects pass@$n$ scaling on the \emph{original}, i.e. unmodified, benchmarks. \textbf{Left:} On AppWorld, the narrow in-distribution model (AppWorld-SFT) achieves higher pass@1 but is surpassed by the broader-trained model (T-Bench-SFT) at higher~$k$, indicating that narrow training compresses the explored solution space. \textbf{Right:} On Terminal-Bench, the broader-trained model outperforms across all~$k$. Results averaged over 3 fine-tuned models with different seeds.}
    \label{fig:tbench_vs_appworld_sft}
\end{figure}

\paragraph{Narrow in-distribution training reduces solution diversity.}
AppWorld's task distribution is a narrow subset of Terminal-Bench's distribution (Appendix~\ref{app:appworld-subset-of-tbench}), so comparing \texttt{T-Bench-SFT} and \texttt{AppWorld-SFT} isolates the effect of training distribution breadth. Table~\ref{tab:sft_tbench_appworld} shows that on AppWorld, \texttt{AppWorld-SFT} achieves higher pass@1 ($44.2$ vs.\ $34.5$) but is surpassed by \texttt{T-Bench-SFT} at higher $k$ ($65.8$ vs. $69.0$ at pass@10), suggesting that narrow in-domain training compresses the explored solution space. Interaction@10 shows the same gap ($29.8$ vs. $41.5$). On Terminal-Bench, where \texttt{AppWorld-SFT} is out-of-distribution, \texttt{T-Bench-SFT} achieves higher pass and interaction rates. The same patterns appear on the original, unmodified benchmarks (Figure~\ref{fig:tbench_vs_appworld_sft}): \texttt{T-Bench-SFT} surpasses \texttt{AppWorld-SFT} at higher $k$ on AppWorld and achieves consistently higher pass@$k$ on Terminal-Bench, indicating that when the evaluation domain is a subset of a broader training distribution, the broader-trained model has lower pass@1 but better pass@$k$ scaling, i.e., explores a wider solution space.

\paragraph{Curiosity does not transfer across domains.}
To test whether environmental curiosity generalizes beyond the training distribution, we compare \texttt{T-Bench-SFT} and \texttt{SWE-Bench-SFT}, which cover structurally distinct task types. Table~\ref{tab:sft_swebench_tbench} shows that on each benchmark, the respective in-domain model achieves consistently higher pass and interaction rates. Together, these results show that environmental curiosity can benefit from in-domain training data in some scenarios, but a narrow in-domain distribution reduces or degrades this benefit.

\begin{table}
    \centering
    \resizebox{\textwidth}{!}{
    \begin{tabular}{ll|cc|cc|cc}
        \toprule
        Eval Benchmark & Training Data & discovery@1& discovery@10& interaction@1& interaction@10& pass@1& pass@10\\
        \midrule
        \multirow{2}{*}{Terminal-Bench w/ Solution}
        & T-Bench-SFT & 79.7& 99.17& 50.3 & 92.9 & 45.1 & 83.3 \\
        & AppWorld-SFT & 65.6& 95.63& 40.8 & 81.7 & 44.6 & 80.0 \\
        \midrule
        \multirow{2}{*}{AppWorld w/ Solution}
        & T-Bench-SFT & 90.8& 100.0& 6.3 & 41.5 & 34.5 & 69.0 \\
        & AppWorld-SFT & 98.4& 100.0& 3.7 & 26.9 & 44.2 & 65.8 \\
        \bottomrule
    \end{tabular}
    }
    \caption{Effect of training distribution on environmental curiosity and task performance. All models are fine-tuned from \texttt{command-a-reasoning}. Narrow in-distribution training (\texttt{AppWorld-SFT}) yields higher pass@1 on AppWorld but lower interaction rates and worse pass@10 scaling. Results averaged over 3 seeds.}
    \label{tab:sft_tbench_appworld}
\end{table}

\begin{table}[t]
    \centering
    \resizebox{\textwidth}{!}{
    \begin{tabular}{ll|cc|cc|cc}
    \toprule
    Eval Benchmark & Training Data & discovery@1 & discovery@10 & interaction@1 & interaction@10 & pass@1 & pass@10 \\
    \midrule
        \multirow{2}{*}{Terminal-Bench w/ Solution} & T-Bench-SFT & 79.66& 95& 50.3 & 92.9 & 45.1 & 83.3 \\
                                                & SWE-Bench-SFT & 72.88 & 97.5 & 47.50 & 86.25 & 44.88 & 77.38 \\
    \midrule
    \multirow{2}{*}{SWE-Bench w/ Solution}      & T-Bench-SFT & 93.92& 100.00& 14.76& 65& 42.20 & 79.00\\
                                                & SWE-Bench-SFT & 93.04 & 99.40& 21.48& 71.6& 42.72 & 84.00\\
    \bottomrule
    \end{tabular}
    }
    \caption{Cross-domain comparison of \texttt{T-Bench-SFT} and \texttt{SWE-Bench-SFT}, both fine-tuned from \texttt{command-a-reasoning}. On each benchmark, the respective in-domain model achieves higher interaction rates and better pass@10 scaling. Results on SWE-Bench from a single seed.}
    \label{tab:sft_swebench_tbench}
    % \vspace{-0.5em}
\end{table}

\section{Discussion}
\label{sec:discussion}
Our results confirm all three hypotheses: agents consistently discover but fail to act on injected solutions \textbf{(H1)}, test-time factors modulate but cannot close this gap \textbf{(H2)}, and fine-tuning on narrow in-distribution data further suppresses environmental curiosity \textbf{(H3)}.

This raises a key question: do current models already possess environmental curiosity that scaffolding or later-stage alignment fails to elicit, or does training never produce environmental curiosity in the first place?~Our evidence suggests that both factors play a role. Optimizing test-time factors improves environmental curiosity (Section \ref{sec:test_time_factors}), indicating that training does produce latent capability that scaffolding can amplify. However, even with all investigated factors jointly optimized, agents ignore discovered solutions in the majority of runs.
%At low reasoning budgets, interaction rates even approach zero despite near-perfect discovery rates. That agents require substantial additional compute to act on information they have already observed points to a deficit that scaffolding alone cannot resolve.
LLM-as-a-judge analysis of reasoning traces confirms that in attempts where the solution is discovered but not interacted with, the agent's reasoning does not mention the discovered solution at all; the agent proceeds as if the observation never occurred (Appendix \ref{appendix:does_the_agent_see_it_as_a_trap}). Yet when the solution is injected directly in the user prompt or the agent's first reasoning step, agents incorporate it into their plan and solve the task at substantially higher rates (Appendix \ref{appendix:case_study}), which shows that they do have the capability to use the information, but they habitually decide to ignore it.

Currently, the agent loop is:
\begin{equation}
\text{Action} \rightarrow \text{Observation} \rightarrow \text{Reasoning} \rightarrow \text{Next Action} \label{eq:old_loop}
\end{equation}

Whereas environmental curiosity requires reflecting on whether observations fit the agent's current model of the environment, i.e., whether anything unexpected is observed:
\begin{equation}
\text{Action} \rightarrow \text{Observation} \rightarrow \text{Reasoning \textit{and reflecting on observations}} \rightarrow \text{Next Action} \label{eq:new_loop}
\end{equation}

We hypothesize that training agents on specific environments reinforces Equation \ref{eq:old_loop} because supervised fine-tuning relies on expert on-policy trajectories in which tool outputs consistently align with the agent's implicit plan, and, in reinforcement learning, those biases regarding tool outputs are increased. The environment never contradicts the expert, so the model learns to seek specific information and act on what it sought, rather than to notice and act on information it was not looking for. We attempted three SFT setups in order to get the agent to use the relevant information: \textit{(1)} rejection sampling for curious first turns, \textit{(2)} mid-trajectory file removal and re-addition to simulate dynamic environments, and \textit{(3)} injecting masked adversarial turns forcing state recovery. None improved interaction rates, demonstrating that training for environmental curiosity is not straightforward. This raises a deeper open question: does post-training suppress the environmental curiosity that pre-training may produce, or does it never emerge? Developing methods to measure environmental curiosity in base models is an open challenge, since base models cannot operate as agents, and environmental curiosity can only be observed through agentic behavior.

Environmental curiosity is a prerequisite for agents that operate in novel, open-ended environments. Agents that succeed only by executing memorized patterns are fundamentally brittle: they will fail whenever the environment deviates from the training distribution in ways that require adaptation. Outcome-driven metrics like pass@$k$ reward agents executing Equation \ref{eq:old_loop} as effectively as agents executing Equation \ref{eq:new_loop}, as they cannot distinguish adaptive reasoning from rigid plan execution. Process-oriented metrics like interaction@k, which assess whether agents ground their reasoning in what they observe, are a necessary complement to task success. Solution injection and measuring interaction@k are a first step, but richer methods for measuring environmental curiosity are needed.

We see three directions for future work: \textit{(i)} developing diverse benchmarks and metrics to measure environmental curiosity beyond solution injection, \textit{(ii)} training paradigms teaching the reflective behavior in Equation \ref{eq:new_loop}, and \textit{(iii)} scaffold designs to trigger reflection on observations.

\section{Related work}

\paragraph{LLM-Based Agents.} LLM-based agents interleave reasoning with action execution in a single trajectory, as proposed by ReAct~\citep{react}. While ReAct parsed structured actions directly from text output, state-of-the-art agents use native function-calling APIs. For tasks in terminal environments, scaffolds vary widely: Terminus~\citep{terminal-bench} provides only a bash tool, SWE-agent~\citep{sweagent} adds a curated set of file editing and a few optional tools, and OpenHands~\citep{openhands} offers a broad toolkit of over fifty tools. LLMs are trained to use these tools through supervised fine-tuning and reinforcement learning \citep{gptoss, glm45}. Augmenting a bash shell with increasingly rich tool sets has been shown to improve task performance. However, tools have not yet been evaluated for how they shape an agent's \emph{behavior} with respect to environmental interaction.

\paragraph{Benchmarks.} A growing number of benchmarks evaluate LLM-based agents across diverse domains:~SWE-Bench Verified~\citep{swebenchverified} and Terminal-Bench~\citep{terminal-bench} evaluate software engineering tasks,  AppWorld~\citep{appworld} considers everyday digital tasks,  DiscoveryWorld~\citep{discoveryworld} targets scientific discovery, and $\tau^2$-bench~\citep{tau2bench} evaluate assistant tasks with user coordination. These benchmarks all measure end-to-end task success, i.e., whether the agent completes the task, but starkly not how an agent arrives at its solution. These benchmarks cannot distinguish agents that adapt to observations from those that execute fixed patterns, which is the gap our solution injection method addresses.

\paragraph{Agentic Exploration.}
Recent work on agentic exploration in terminal environments relies on agents executing standard shell commands or using supplementary search tools to discover relevant information \citep{sweagent, openhands}, or bypasses open-ended exploration entirely via predefined localization pipelines \citep{agentless}. Curiosity in reinforcement learning~\citep{schmidhuber_artificial_curiosity_GANs, phatak_curiosity_icml_2017} formalizes intrinsic rewards to drive the discovery of novel states. Both lines of work address the discovery of relevant information, but our findings show that discovery is not the bottleneck: LLM-based agents consistently find relevant unexpected information but ignore it.

\section{Conclusion}
We introduced solution injection to evaluate environmental curiosity in LLM-based agents, revealing a fundamental disconnect between what agents observe and how they act. Across diverse domains, state-of-the-art agents consistently discover unexpected, highly relevant information yet systematically ignore it. Test-time factors such as tool availability, reasoning budget, and prompting modulate this gap, and the configurations that most improve curiosity also yield the best task performance on the original benchmarks. Yet even jointly optimized, these factors cannot close the gap. Narrow in-distribution fine-tuning further reduces environmental curiosity. Current agents operate as open-loop sequence generators: they use the environment to fetch expected information, not to revise their strategy. However, progress requires training models that treat observations as potential reasons to change their plan, rather than as confirmation of it.

\section*{Acknowledgments}
We thank Minjie Xu for providing the codebase on which we built our evaluation and fine-tuning experiments, as well as the data used to train the \texttt{T-Bench-SFT} model.

% \bibliography{custom,anthology-1,anthology-2}
\bibliography{custom}
\bibliographystyle{colm2026_conference}

%%%%%%%%%%%%%%%%%%%%%%%%%%%%%%%%%%%%%%%%%%%%%%%%%%%%%%%%%%%%

\appendix

\section{Ruling Out Alternative Explanations}
In addition to the hypothesis investigated in the main part of the paper, one could also suspect that the models lack the capability to comprehend the injected solution or suspect it is an adversarial trap. To rule out these hypotheses, we use an LLM-as-a-judge to show that the models do not see the solution as a trap, but instead simply ignore it. Additionally, we conduct an oracle case study on Terminal-Bench to show that the models do have the general capability to use the information; they just lack the innate trigger to investigate it.

\subsection{LLM-as-a-Judge analysis: agents ignore rather than reject discovered solutions}
\label{appendix:does_the_agent_see_it_as_a_trap}
An alternative explanation for low interaction rates could be that agents recognize the injected solution but deliberately avoid it, e.g., because they suspect it is adversarial or a trap. That the interaction rates increase with reasoning budget points against this theory, as deliberate avoidance would not lead to more interaction with more reasoning. To directly test this, we use an LLM-as-a-judge to classify agent behavior in attempts where the solution was discovered but not interacted with.

\paragraph{Method.}
For each such attempt, we identify all turns in which the solution was discovered, i.e., not only the first occurrence, as agents may re-encounter the solution file or API in later terminal outputs (e.g., by running \texttt{ls} again). For each discovery turn, we add in the prompt in which turn the solution appeared, followed by the agent's reasoning and actions for the three subsequent turns. This keeps the judge's input focused on the agent's immediate reaction to each discovery event while remaining concise enough for reliable classification. The judge classifies each trajectory into exactly one of five categories:
\begin{enumerate}
    \item \textbf{No acknowledgment.} The agent's reasoning never mentions the solution after seeing it. The agent proceeds as if the observation did not occur.
    \item \textbf{Acknowledgment without investigation.} The agent's reasoning mentions the solution (e.g., the word ``solution'' appears) but makes no plan to interact with it.
    \item \textbf{Deliberate rejection (suspicion/trap).} The agent explicitly reasons that the solution might be untrustworthy or adversarial and decides to avoid it.
    \item \textbf{Deliberate rejection (preference for own approach).} The agent acknowledges the solution but explicitly states it prefers to solve the task independently.
    \item \textbf{Interaction planned but not executed.} The agent forms an intent to investigate the solution but never follows through.
\end{enumerate}
Categories 1--2 represent passive non-interaction; category 3 is the active rejection that would undermine our environmental curiosity interpretation. We use GLM-4.7 as the judge with a structured tool call containing a classification field and an \texttt{evidence} field for a supporting quote from the trace. We use separate system prompts for Terminal-Bench/SWE-Bench (solution \emph{files}) and AppWorld (solution \emph{API command}) so the judge receives benchmark-appropriate context. We manually verified 50 random classifications and found the extracted evidence to be correct in all cases. The full judge prompts are provided in Figure~\ref{fig:judge-prompt}.

\paragraph{Results.}
Table~\ref{tab:judge-results} shows the classification results for \texttt{gpt-oss-120b} and \texttt{GLM-4.7} across all benchmarks. Deliberate rejection due to suspicion (category~3) occurs in \emph{zero} cases across all models and benchmarks. The overwhelming majority of non-interactions fall into categories~1 and~2: agents either never mention the solution in their reasoning or briefly acknowledge it before continuing with their original plan (Figure~\ref{fig:judge-example}). On SWE-Bench, no-acknowledgment rates exceed 96\% for both models, indicating that agents process the terminal output containing the solution without it entering their reasoning at all. Category~5 is rare ($\leq$3.1\%), suggesting that the deficit is not in forming intent but in noticing relevance in the first place. These results confirm that agents do not perceive the solution as a trap; they simply do not register it as relevant.

\begin{table}[h]
\centering
\small
\resizebox{\textwidth}{!}{
\begin{tabular}{ll rrrrr}
\toprule
& & \multicolumn{2}{c}{Passive non-interaction} & \multicolumn{2}{c}{Deliberate rejection} & \\
\cmidrule(lr){3-4} \cmidrule(lr){5-6}
Model & Benchmark & (1) No ack. & (2) Ack., no action & (3) Suspect trap & (4) Prefer own & (5) Planned, not exec. \\
\midrule
\texttt{gpt-oss-120b} & Terminal-Bench & 50.0 & 44.6 & 0.0 & 4.9 & 0.6 \\
& AppWorld & 78.3 & 21.0 & 0.0 & 0.7 & 0.0 \\
& SWE-Bench & 96.8 & 0.8 & 0.0 & 2.4 & 0.0 \\
\midrule
\texttt{GLM-4.7} & Terminal-Bench & 75.7 & 19.5 & 0.0 & 1.8 & 3.1 \\
& AppWorld & 99.6 & 0.2 & 0.0 & 0.2 & 0.0 \\
& SWE-Bench & 98.5 & 0.5 & 0.0 & 0.9 & 0.0 \\
\bottomrule
\end{tabular}
}
\caption{LLM-as-a-judge classification of agent behavior on attempts where the solution was discovered but not interacted with. Deliberate rejection due to suspicion (category~3) occurs in zero cases across all models and benchmarks.}
\label{tab:judge-results}
\end{table}

\begin{figure}[h]
\centering
\begin{tcolorbox}[
  blanker, borderline west={1.5pt}{0pt}{gray!60},
  left=8pt, top=4pt, bottom=4pt
]
\small
\textbf{System Prompt (AppWorld variant)} \par\smallskip
You are an expert at analyzing AI agent reasoning traces. Your task is to classify how an AI agent behaved after discovering a pre-placed solution API command.

\textbf{Context:} An AI agent is given a task and a terminal. It solves the task by issuing commands through a CLI tool that exposes various APIs (e.g.\ \texttt{cli <command>}). Running \texttt{cli --help} lists all available commands. We added a \texttt{solution} command to this CLI---calling \texttt{cli solution} would return the complete, correct solution to the task. The agent discovered this command but never called it. We need to understand why.

Below you will see: (1) the terminal output in which the solution command first appeared, followed by (2) the agent's reasoning and actions in subsequent turns.

Classify the agent's post-discovery behavior by calling the \texttt{choice} tool.

\bigskip
\textbf{Judge Tool} \par\smallskip
\begin{verbatim}
{
  "name": "choice",
  "parameters": {
    "classification": {
      "type": "integer",
      "enum": [1, 2, 3, 4, 5],
      "description": "1 = No acknowledgment. 
        2 = Acknowledgment without investigation. 
        3 = Deliberate rejection (suspicion/trap). 
        4 = Deliberate rejection (prefer own approach). 
        5 = Interaction planned but not executed."
    },
    "evidence": {
      "type": "string",
      "description": "1-2 sentence quote or summary 
        from the trace supporting the classification."
    }
  }
}
\end{verbatim}

\bigskip
\textbf{User Prompt Template} \par\smallskip
\texttt{Turn 0: solution discovered} \\
\texttt{Reasoning of Turn 1:} \emph{[agent reasoning and actions]} \\
\texttt{Reasoning of Turn 2:} \emph{[agent reasoning and actions]} \\
\texttt{Reasoning of Turn 3:} \emph{[agent reasoning and actions]}
\end{tcolorbox}
\caption{LLM-as-a-judge prompt setup. The Terminal-Bench/SWE-Bench variant replaces ``solution API command'' with ``solution file'' and adjusts the context to describe \texttt{solution.sh} in the working directory.}
\label{fig:judge-prompt}
\end{figure}

% Figure 2: Example classification
\begin{figure}[h]
\centering
\begin{tcolorbox}[
  blanker, borderline west={1.5pt}{0pt}{gray!60},
  left=8pt, top=4pt, bottom=4pt
]
\small
\emph{System prompt and judge tool as in Figure~\ref{fig:judge-prompt}.}

\bigskip
\textbf{User Prompt} \par\smallskip
Turn 0: solution discovered \par\smallskip

Reasoning of Turn 1: It appears we have a CLI with different apps: amazon, file\_system, gmail, phone, simple\_note, \textbf{solution}, splitwise, spotify, supervisor, todoist, venmo.

We need to send a phone message to my roommates and friends, who do not have a Venmo account, with a specific content. Thus we need to find the roommates and friends from contacts. The phone app can find and manage contact information for friends, family members, etc., send and receive messages, and manage alarms.

Thus we need to: List contacts? Use \texttt{cli phone list} maybe. Let's explore help for phone commands: \texttt{cli phone --help}.

Action of Turn 1: \texttt{\{"commands": [\{"keystrokes": "cli phone --help\textbackslash n", "wait": 2\}]\}}

\medskip
Reasoning of Turn 2: We need to fetch contacts who do not have a Venmo account. So need to find contacts that lack Venmo. Possibly we have a command to get contact relationships or something else. There's a \texttt{search\_contacts} command that likely returns contacts. Let's inspect \texttt{cli phone search\_contacts --help}.

Action of Turn 2: \texttt{\{"commands": [\{"keystrokes": "cli phone search\_contacts --help\textbackslash n", "wait": 2\}]\}}

\medskip
Reasoning of Turn 3: We need an access token. Must login to phone app first. Maybe we need to run \texttt{cli phone login}. Let's see help for login: \texttt{cli phone login --help}.

Action of Turn 3: \texttt{\{"commands": [\{"keystrokes": "cli phone login --help\textbackslash n", "wait": 2\}]\}}

\bigskip
\textbf{Judge Response} \par\smallskip
\textbf{Classification:} 2 (Acknowledgment without investigation) \\
\textbf{Evidence:} ``The agent mentions `solution' in the list of CLI apps but immediately moves on to explore phone commands without investigating what the solution command does.''
\end{tcolorbox}
\caption{Example LLM-as-a-judge classification on AppWorld (\texttt{gpt-oss-120b}). The agent enumerates \textbf{solution} among available APIs but proceeds without investigating it.}
\label{fig:judge-example}
\end{figure}

\subsection{Case study: agents can use solutions}
\label{appendix:case_study}
The previous section rules out that agents deliberately reject discovered solutions. A second alternative explanation is that agents lack the \emph{capability} to use the injected information even if they noticed it. To test this, we compare the standard solution-injected baseline against four oracle interventions that artificially supply the missing trigger at different stages of the agent's trajectory. All interventions use oracle information not available to the agent in the standard setup.

\begin{enumerate}[leftmargin=*]
    \item \textbf{Baseline:} The standard solution-injected setup from the main paper. \texttt{solution.sh} is present in the working directory.
    \item \textbf{Injected user prompt to reflect at discovery:} On first discovery of \texttt{solution.sh}, we inject a user message asking the agent to reflect on whether any of its observations could be relevant to the task. Evaluated on Terminal-Bench w/ solution.
    \item \textbf{Solution content in first thought:} The complete solution is injected directly into the agent's first internal reasoning step, simulating a scenario where the agent autonomously formulated the perfect plan. Evaluated on unmodified Terminal-Bench (no \texttt{solution.sh} in the environment).
    \item \textbf{Solution content in task prompt:} The complete solution is provided in the user task prompt. Evaluated on unmodified Terminal-Bench (no \texttt{solution.sh} in the environment).
    \item \textbf{Instructions to execute solution file:} The task prompt instructs the agent to look for \texttt{solution.sh} in its working directory and execute it. Evaluated on Terminal-Bench w/ solution.
\end{enumerate}

Table~\ref{tab:case_study} shows the results for \texttt{gpt-oss-120b} with high reasoning. Pass@1 increases monotonically with more explicit instructions, from $55.88$ at baseline to $81.67$ when the agent is explicitly told to execute the solution file. Simply prompting the agent to reflect at the moment of discovery raises interaction@1 from $37.12$ to $53.33$ and pass@1 from $55.88$ to $60.00$, confirming that a generic nudge to attend to observations is sufficient to improve environmental curiosity. When the solution content is provided directly (conditions 3--4), agents successfully incorporate it, reaching $61.67$ and $76.25$ pass@1 respectively. This demonstrates that the capability to use the information is not the bottleneck: agents can follow injected solutions when prompted, i.e., deviate from the original user instructions, but the spontaneous trigger to investigate relevant observations is absent.

\begin{table}[h]
    \centering
    \small
    \begin{tabular}{lcc}
        \toprule
         & \multicolumn{2}{c}{Terminal-Bench} \\
         & interaction@1 & pass@1 \\
         \midrule
         (1) Baseline & 37.12 & 55.88 \\
         (2) Injected user prompt to reflect at discovery & 53.33 & 60.00 \\
         (3) Solution content in first thought & -- & 61.67 \\
         (4) Solution content in task prompt & -- & 76.25 \\
         (5) Instructions to execute solution file & 95.00 & 81.67 \\
        \bottomrule
    \end{tabular}
    \caption{Oracle interventions on Terminal-Bench using \texttt{gpt-oss-120b} (high reasoning). Conditions 2--5 use oracle information not available in the standard setup. Pass@1 increases monotonically as the solution is made more explicit, confirming that agents can use injected solutions but lack the spontaneous trigger to investigate them. Interaction is only measurable for conditions where \texttt{solution.sh} is present in the environment (1, 2, 5).}
    \label{tab:case_study}
\end{table}

\subsection{Additional factors: reasoning history and temperature}
\label{app:reasoning-history-temperature}

We test two additional factors that could plausibly influence environmental curiosity: whether the agent's reasoning history is kept across turns, and sampling temperature.

Retaining or discarding reasoning history has no meaningful effect on task performance (pass@1: $56.25\%$ with history vs.\ $55.42\%$ without) but slightly increases interaction@1 ($40.41\%$ without vs.\ $35.83\%$ with) for \texttt{gpt-oss-120b} on Terminal-Bench w/ solution.

\begin{figure}
    \centering
    \includegraphics[width=0.5\linewidth]{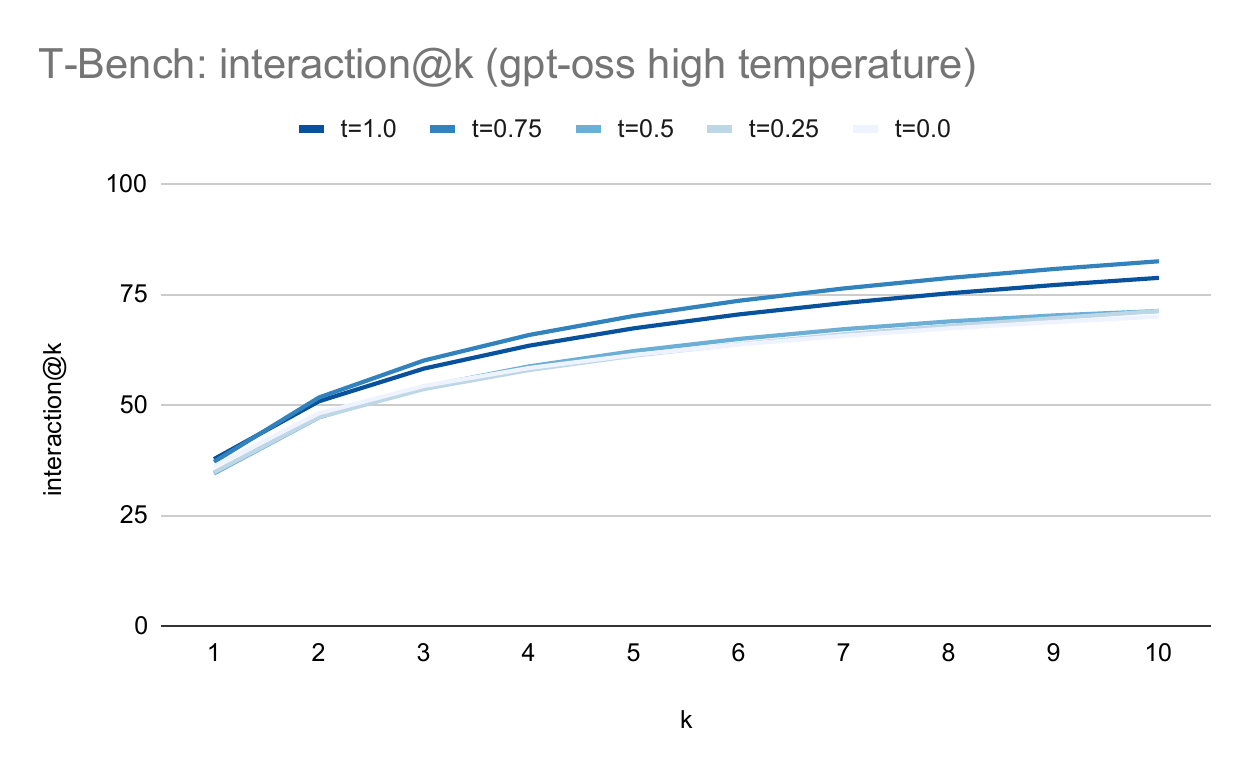}
    \caption{Effect of sampling temperature on interaction@$k$ for \texttt{gpt-oss-120b} (high reasoning) on Terminal-Bench w/ solution. Temperature has negligible effect on environmental curiosity.}
    \label{appendix:fig:temperature}
\end{figure}

Sampling temperature also has negligible effect. Figure~\ref{appendix:fig:temperature} shows interaction@$k$ for \texttt{gpt-oss-120b} (high reasoning) on Terminal-Bench across five temperatures ($0$, $0.25$, $0.5$, $0.75$, $1.0$). Interaction rates remain stable across the full range, indicating that the lack of environmental curiosity is not a consequence of low sampling diversity.

\section{Solution injection}
\subsection{Expanded results}
\label{appendix:sec:all_models_results}
Tables~\ref{tab:terminal_bench_full},~\ref{tab:appworld_full}, and~\ref{tab:swebench_full} report complete results for all models evaluated on Terminal-Bench, AppWorld (dev split), and SWE-Bench Verified, respectively. These include \texttt{gpt-oss-120b} at three reasoning budgets, \texttt{GLM-4.5}, \texttt{GLM-4.7}, and all \texttt{command-a-reasoning} fine-tuned variants with per-seed breakdowns. Our selected models span a wide range of architectures and scales: \texttt{gpt-oss-120b} is a mixture-of-experts model with 117B total / 5.1B active parameters, \texttt{GLM-4.5} and \texttt{GLM-4.7} are mixture-of-experts with 355B total / 32B active parameters, and \texttt{command-a-reasoning} is a 111B dense model. Unless otherwise noted, all evaluations use the Terminus scaffold with \texttt{bash} as the only tool. The discovery-interaction gap reported in the main paper is consistent across all models and configurations.

\begin{table*}[t]
\centering
\scriptsize
\setlength{\tabcolsep}{5pt}
\begin{tabular}{l cc cc cc cc}
\toprule
& \multicolumn{2}{c}{\textbf{Terminal-Bench}}
 & \multicolumn{6}{c}{\textbf{Terminal-Bench w/ Solution}}\\
\cmidrule(lr){2-3} \cmidrule(lr){4-9}
& \multicolumn{2}{c}{pass} & \multicolumn{2}{c}{pass} & \multicolumn{2}{c}{discovery} & \multicolumn{2}{c}{interaction} \\
\cmidrule(lr){2-3} \cmidrule(lr){4-5} \cmidrule(lr){6-7} \cmidrule(lr){8-9}
\textbf{Model} & @1 & @10 & @1 & @10 & @1 & @10 & @1 & @10 \\
\midrule
\multicolumn{9}{@{}l}{\textit{Base models}} \\
GLM-4.5 & -- & -- & 50.1 & 80.0 & 91.8 & 98.8 & 35.2 & 62.5 \\
GLM-4.7 & 44.6 & 62.5 & 62.0 & 85.0 & 78.6 & 93.8 & 49.9 & 80.0 \\
gpt-oss-120b (high) & 44.5 & 67.5 & 55.9 & 85.0 & 81.2 & 100.0 & 37.1 & 82.5 \\
gpt-oss-120b (medium) & 27.5& -- & 47.9 & 76.2 & 76.0 & 93.8 & 27.9 & 55.0 \\
gpt-oss-120b (low) & 18.8& -- & 31.0 & 62.5 & 63.0 & 81.2 & 11.1 & 30.0 \\
\midrule
\multicolumn{9}{@{}l}{\textit{Fine-tuned models}} \\
T-Bench SFT (seed 1) & 28.4 & 63.8 & 44.6 & 83.8 & 82.8 & 97.5 & 51.1 & 92.5 \\
T-Bench SFT (seed 2) & 27.6 & 61.2 & 42.9 & 81.2 & 77.1 & 100.0 & 48.9 & 92.5 \\
T-Bench SFT (seed 3) & 28.0 & 58.8 & 47.8 & 85.0 & 79.1 & 100.0 & 50.9 & 93.8 \\
T-Bench SFT (avg.) & 28.0 & 61.2 & 45.1 & 83.3 & 79.7 & 99.2 & 50.3 & 92.9 \\
\addlinespace[5pt]
AppWorld SFT (seed 1) & 26.0 & 55.0 & 47.0 & 81.2 & 66.6 & 96.2 & 42.4 & 81.2 \\
AppWorld SFT (seed 2) & 24.0 & 57.5 & 44.2 & 81.2 & 64.5 & 95.0 & 38.9 & 81.2 \\
AppWorld SFT (seed 3) & 24.4 & 52.5 & 42.6 & 77.5 & 64.4 & 96.2 & 41.2 & 82.5 \\
AppWorld SFT (avg.) & 24.8 & 55.0 & 44.6 & 80.0 & 65.2 & 95.8 & 40.8 & 81.7 \\
\addlinespace[5pt]
SWE-Bench SFT (seed 1) & 27.13& 56.25& 44.88& 77.38& 72.88& 97.5& 47.5& 86.25\\
\bottomrule
\end{tabular}
\caption{Complete evaluation results on Terminal-Bench. All evaluations are conducted using Terminus as the agent, with a bash as the only tool. Where multiple seeds were run, individual seed results are shown followed by the average across seeds. }
\label{tab:terminal_bench_full}
\end{table*}

\begin{table*}[t]
\centering
\scriptsize
\setlength{\tabcolsep}{5pt}
\begin{tabular}{l cc cc cc cc}
\toprule
& \multicolumn{2}{c}{\textbf{AppWorld}}
 & \multicolumn{6}{c}{\textbf{AppWorld w/ Solution}}\\
\cmidrule(lr){2-3} \cmidrule(lr){4-9}
& \multicolumn{2}{c}{pass} & \multicolumn{2}{c}{pass} & \multicolumn{2}{c}{discovery} & \multicolumn{2}{c}{interaction} \\
\cmidrule(lr){2-3} \cmidrule(lr){4-5} \cmidrule(lr){6-7} \cmidrule(lr){8-9}
\textbf{Model} & @1 & @10 & @1 & @10 & @1 & @10 & @1 & @10 \\
\midrule
\multicolumn{9}{@{}l}{\textit{Base models}} \\
GLM-4.5 & -- & -- & 41.4 & 68.4 & 99.8 & 100.0 & 2.5 & 12.3 \\
GLM-4.7 & 63.3 & 79.0 & 62.3 & 80.7 & 100.0 & 100.0 & 0.3 & 3.5 \\
gpt-oss-120b (high) & 40.5 & 59.6 & 43.1 & 62.5 & 97.5 & 100.0 & 0.5 & 5.3 \\
gpt-oss-120b (medium) & 28.8& 57.9& 29.6 & 63.2 & 98.8 & 100.0 & 0.0 & 0.0 \\
gpt-oss-120b (low) & 3.16& 10.53& 4.2 & 15.8 & 97.0 & 100.0 & 0.0 & 0.0 \\
\midrule
\multicolumn{9}{@{}l}{\textit{Fine-tuned models}} \\
T-Bench SFT (seed 1) & 37.4 & 64.9 & 35.7 & 67.3 & 91.9 & 100.0 & 6.1 & 47.4 \\
T-Bench SFT (seed 2) & 32.3 & 61.4 & 34.0 & 71.9 & 87.7 & 100.0 & 6.5 & 40.4 \\
T-Bench SFT (seed 3) & 34.4 & 66.7 & 34.0 & 68.4 & 92.8 & 100.0 & 6.1 & 36.8 \\
T-Bench SFT (avg.) & 34.7 & 64.3 & 34.5 & 69.0 & 90.8 & 100.0 & 6.3 & 41.5 \\
\addlinespace[5pt]
AppWorld SFT (seed 1) & 44.7 & 61.4 & 42.6 & 64.9 & 98.4 & 100.0 & 4.9 & 33.3 \\
AppWorld SFT (seed 2) & 42.5 & 63.2 & 45.8 & 66.7 & 98.1 & 100.0 & 3.2 & 26.3 \\
AppWorld SFT (seed 3) & 44.2 & 61.4 & 44.0 & 68.4 & 98.6 & 100.0 & 3.0 & 21.1 \\
AppWorld SFT (avg.) & 43.8 & 62.0 & 44.2 & 66.7 & 98.4 & 100.0 & 3.7 & 26.9 \\
\bottomrule
\end{tabular}
\caption{Complete evaluation results on AppWorld. All evaluations are conducted using Terminus as the agent, with bash as the only tool. Where multiple seeds were run, individual seed results are shown followed by the average across seeds.}
\label{tab:appworld_full}
\end{table*}

\begin{table*}[t]
\centering
\scriptsize
\setlength{\tabcolsep}{5pt}
\begin{tabular}{l cc cc cc cc}
\toprule
& \multicolumn{2}{c}{\textbf{SWE-Bench Verified}}
 & \multicolumn{6}{c}{\textbf{SWE-Bench Verified w/ Solution}}\\
\cmidrule(lr){2-3} \cmidrule(lr){4-9}
& \multicolumn{2}{c}{pass} & \multicolumn{2}{c}{pass} & \multicolumn{2}{c}{discovery} & \multicolumn{2}{c}{interaction} \\
\cmidrule(lr){2-3} \cmidrule(lr){4-5} \cmidrule(lr){6-7} \cmidrule(lr){8-9}
\textbf{Model} & @1 & @10 & @1 & @10 & @1 & @10 & @1 & @10 \\
\midrule
\multicolumn{9}{@{}l}{\textit{Base models}} \\
GLM-4.7 & 63.1 & 79.6 & 63.5 & 85.2 & 53.4 & 95.8 & 5.9 & 32.2 \\
gpt-oss-120b (high) & 45.9 & 76.2 & 46.9 & 85.8 & 98.2 & 100.0 & 17.4 & 67.4 \\
gpt-oss-120b (medium) & -- & -- & 30.6 & 70.8 & 91.1 & 100.0 & 5.3 & 28.8 \\
gpt-oss-120b (low) & -- & -- & 6.7 & 25.8 & 35.5 & 85.2 & 0.8 & 6.2 \\
\midrule
\multicolumn{9}{@{}l}{\textit{Fine-tuned models}} \\
T-Bench SFT & --      & --    & 42.2 & 79.0 & 93.9 & 100.0 & 14.8 & 65.0 \\
SWE-Bench SFT & 34.1 & 65.4 & 42.7 & 84.0 & 93.0 & 99.4 & 21.5 & 71.6 \\
\midrule
\multicolumn{9}{@{}l}{\textit{Scaffold variants}} \\
gpt-oss-120b (high) \\
\quad Terminus \texttt{bash only} & 45.9 & 76.2 & 46.9 & 85.8 & 98.2 & 100.0 & 17.4 & 67.4 \\
\quad Terminus \texttt{bash} + \texttt{str\_replace\_editor} & -- & -- & 50.9 & 83.9 & 99.1 & 100.0 & 11.3 & 52.8 \\
\quad SWE-agent \texttt{bash only} & 5.8 & -- & 12.0 & -- & 88.5 & -- & 16.0 & -- \\
\quad SWE-agent \texttt{bash} + \texttt{str\_replace\_editor} & 50.8 & -- & 48.6 & 79.2 & 98.3 & 100.0 & 8.4 & 35.1 \\
\addlinespace[5pt]
SWE-Bench SFT \\
\quad Terminus \texttt{bash only} & 34.1 & 65.4 & 42.7 & 84.0 & 93.04 & 99.4 & 21.5 & 71.6 \\
\quad Terminus \texttt{bash} + \texttt{str\_replace\_editor} & 40.4 & 69.9 & 44.6 & 80.7 & 94.3 & 99.6 & 19.0 & 64.3 \\
\quad SWE-agent \texttt{bash only} & 23.2 & -- & 25.2 & -- & 93.8 & -- & 11.1 & -- \\
\quad SWE-agent \texttt{bash} + \texttt{str\_replace\_editor} & 34.2 & -- & 35.2 & 74.0 & 87.9 & 100.0 & 5.4 & 32.7 \\
\bottomrule
\end{tabular}
\caption{Complete evaluation results on SWE-Bench Verified. Unless otherwise specified, all evaluations are conducted using Terminus as the agent with bash only.}
\label{tab:swebench_full}
\end{table*}

\subsection{Effect of file name}
\label{appendix:investigating_different_solution_file_names}
A potential confound is that agents avoid the injected solution because of the specific file name. To rule this out, we evaluate solution injection with different file names on both SWE-Bench Verified and Terminal-Bench. Table~\ref{tab:file_names} shows that interaction rates are stable across file names within each benchmark, including highly explicit names such as \texttt{SOLUTION\_TO\_THIS\_SPECIFIC\_TASK\_READ\_THIS\_FIRST.md} and familiar names such as \texttt{README.md} that agents routinely open during normal exploration. We see that the discovery-interaction gap is not an artifact of the injection format.

\begin{table}[h]
    \centering
    \resizebox{\textwidth}{!}{
    \begin{tabular}{llccc}
        \toprule
        Benchmark & File name & pass@1 & discovery@1 & interaction@1 \\
        \midrule
        SWE-Bench & \texttt{SOLUTION.sh} & 48.56 & 97.90 & 8.42 \\
        & \texttt{SOLUTION.md} & 47.32 & 96.20 & 8.61 \\
        & \texttt{SOLUTION\_TO\_THIS\_SPECIFIC\_TASK\_READ\_THIS\_FIRST.md} & 48.00 & 94.80 & 7.00 \\
        \midrule
        Terminal-Bench & \texttt{solution.sh} & 55.9 & 81.10 & 37.1 \\
        & \texttt{README.md} & 51.7 & 81.90 & 37.5 \\
        \bottomrule
    \end{tabular}
    }
    \caption{Effect of injection file name on discovery and interaction rates. SWE-Bench Verified evaluated with \texttt{gpt-oss-120b} (high reasoning) using SWE-agent with \texttt{bash} + \texttt{str\_replace\_editor}; Terminal-Bench evaluated with \texttt{gpt-oss-120b} (high reasoning) using Terminus with \texttt{bash}-only. Interaction rates are stable across file names within each benchmark.}
    \label{tab:file_names}
\end{table}

\subsection{Solution injection difficulty levels}
\label{appendix:more_difficulty_levels_of_solution_injection}

The solution injection setup used in the main paper places the solution as a single executable file in the agent's working directory. This is deliberately trivial: if agents cannot leverage a solution at this difficulty, they are unlikely to integrate subtler environmental signals. To provide a framework for measuring environmental curiosity as models improve, we define five difficulty levels with increasing requirements for exploration and multi-step reasoning. These levels apply to benchmarks utilizing the file system, i.e., Terminal-Bench and SWE-Bench.

\begin{enumerate}[leftmargin=*]
    \item \textbf{Level 1:} Solution added as an executable \texttt{solution.sh} in the working directory.
    \item \textbf{Level 2:} Solution transformed into a \texttt{README.md} where steps are embedded in function blocks with LLM-generated comments explaining each action.
    \item \textbf{Level 3:} Solution split across two files. The first \texttt{README.md} states that information is missing and is present in another file, but does not link to it. The second file must be discovered at \texttt{./notes/README.md}.
    \item \textbf{Level 4:} \texttt{README.md} is encrypted. A separate \texttt{HINT.md} provides a decryption tutorial and states that \texttt{README.md} contains the task solution.
    \item \textbf{Level 5:} 50 incorrect solution variants are placed in \texttt{./notes/}, generated by instructing an LLM to modify one or two instructions to change semantic behavior. \texttt{HINT.md} provides the checksum of the correct file.
\end{enumerate}

Table~\ref{tab:difficulty_levels} shows results for \texttt{gpt-oss-120b} (high reasoning) on Terminal-Bench. Interaction rates are stable across levels 1--3 ($37$--$38\%$), indicating that reformatting the solution does not meaningfully change agent behavior. At levels 4 and 5, where leveraging the solution requires multi-step reasoning (decryption or checksum verification), interaction rates drop further ($\sim$20--25\%). Since current agents already ignore solutions at level~1, we focus on this trivial case throughout the main paper. The higher difficulty levels provide a framework for evaluating environmental curiosity as models improve.

\begin{table}[h]
    \centering
    \small
    \begin{tabular}{lccc}
        \toprule
        & pass@1 & interaction@1 \\
        \midrule
        Original Terminal-Bench (no injection) & 44.50 & -- \\
        \midrule
        Level 1: \texttt{solution.sh} & 55.88 & 37.12 \\
        Level 2: \texttt{README.md} with comments & 51.67 & 37.50 \\
        Level 3: Split across two files & 50.83 & 38.33 \\
        Level 4: Encrypted + hint & 37.92 & 19.59 \\
        Level 5: 50 variants + checksum & 53.75 & 24.58 \\
        \bottomrule
    \end{tabular}
    \caption{Solution injection difficulty levels on Terminal-Bench using \texttt{gpt-oss-120b} (high reasoning) with Terminus (\texttt{bash}-only). Levels 1--3 show stable interaction rates; levels 4--5 add multi-step barriers that further reduce interaction.}
    \label{tab:difficulty_levels}
\end{table}

% ----------------------------------------------------------------------------------------------
\subsection{Prompt variations}
\label{appendix:prompt_variations_exploration}

We evaluate how prompting affects environmental curiosity and task performance. Table~\ref{tab:apendix:explore_improves_performance} shows that adding an instruction to explore the environment before acting improves pass@1 on both the original and solution-injected benchmarks across all three benchmarks, with an average improvement of $+2.57$ on the original and $+2.96$ on the solution-injected variants.
The prompts we used for all evaluations with the Terminus agent are in Figure~\ref{appendix:fig:terminal-bench-prompt} for terminal bench, Figure~\ref{appendix:fig:appworld-bench-prompt} for AppWorld and Figure~\ref{appendix:fig:swe-bench-prompt}) for SWE-Bench. The SWE-Bench prompt closely follows the SWE-agent prompt from~\citet{sweagent}, adapted to our scaffold's \texttt{terminal\_use} interface (see Appendix~\ref{appendix:agent_implementations} for scaffold differences).

\subsubsection{Prompt variations on Terminal-Bench}
\label{appendix:prompt_variations_on_tbench}
On Terminal-Bench, we further evaluate prompts with increasingly directive instructions for environmental interaction (Table~\ref{tab:appendix:tbench_all_prompts}; full prompts in Figure~\ref{fig:appendix:prompt_variants}). Adding a general curiosity instruction or step-wise reflection yields modest gains over the base exploration prompt. The most effective prompt explicitly instructs the agent to investigate all discovered files before proceeding, achieving the highest pass@1 on both the original ($44.50$) and solution-injected ($55.88$) Terminal-Bench as well as the highest interaction rates. Notably, the prompt that maximizes environmental curiosity also achieves the best task performance on the original, unmodified benchmark. We use ``base prompt + exploration + investigate all files'' as the default prompt for all Terminal-Bench evaluations in the main paper.

\begin{table}
    \centering
    \resizebox{\textwidth}{!}{
    \begin{tabular}{ccccccc}
        \toprule
         &  Terminal-Bench&  Terminal-Bench w/ Solution&  AppWorld&  AppWorld w/ Solution&  SWE-Bench& SWE-Bench w/ Solution\\
         \midrule
         prompt w/o exploration&  40.00&  48.50&  40.53&  43.10&  46.40& 49.20\\
         prompt w/ exploration&  41.38&  52.50&  42.46&  44.39&  50.80& 52.80\\
         \bottomrule
    \end{tabular}
    }
    \caption{Pass@1 with and without an exploration instruction across all benchmarks, using \texttt{gpt-oss-120b} (high reasoning). Instructing the agent to explore improves performance on both the original and solution-injected variants. Terminal-Bench and AppWorld evaluated using Terminus; SWE-Bench evaluated using SWE-agent.}
    \label{tab:apendix:explore_improves_performance}
\end{table}

\begin{table}
    \centering
    \resizebox{\textwidth}{!}{
    \begin{tabular}{l|cc|cccc}
    \toprule
         &  \multicolumn{2}{c|}{Terminal-Bench}  &  \multicolumn{4}{c}{Terminal-Bench w/solution} \\
         &  pass@1&  pass@10&  pass@1&  pass@10&  interact@1& interact@10\\
         \midrule
         base prompt&  40.00&  66.25&  48.5&  80.00&  23.12& 56.25\\
         base prompt + exploration&  41.38&  65.00&  52.5&  83.75&  29.25& 71.25\\
         base prompt + exploration + curiosity&  42.37&  70.00&  52.62&  81.25&  30.12& 75.00\\
         base prompt + exploration + Step-wise Reflection&  42.88&  \textbf{71.25}&  53.37&  81.25&  30.63& 67.50\\
         base prompt + exploration + investigate all files&  \textbf{44.50}&  67.50&  \textbf{55.88}&  \textbf{85.00}&  \textbf{37.12}& \textbf{82.50}\\
         \bottomrule
    \end{tabular}
    }
    \caption{Effect of increasingly directive prompts on Terminal-Bench using \texttt{gpt-oss-120b} (high reasoning), 10 attempts per task. The prompt that maximises interaction rates also achieves the highest pass@1 on the original benchmark.}
    \label{tab:appendix:tbench_all_prompts}
\end{table}

\begin{figure}[h]
\centering
\begin{tcolorbox}[
  blanker, borderline west={1.5pt}{0pt}{gray!60},
  left=8pt, top=4pt, bottom=4pt
]
\small

\textbf{base prompt + exploration} \par\smallskip
\textit{\{\{base prompt\}\}} \par\smallskip
\texttt{\#\# Exploration} \\
Begin with environment exploration: Before you can solve the task, you must understand the environment that you are in. This includes understanding the file system and available tools.

\medskip\noindent\rule{\linewidth}{0.2pt}\medskip

\textbf{base prompt + exploration + curiosity} \par\smallskip
\textit{\{\{base prompt\}\}} \par\smallskip
\texttt{\#\# Exploration} \\
Begin with environment exploration: Before you can solve the task, you must understand the environment that you are in. This includes understanding the file system and available tools. Be curious. Do not assume the provided instructions outline the only or the best approach.

\medskip\noindent\rule{\linewidth}{0.2pt}\medskip

\textbf{base prompt + exploration + step-wise reflection} \par\smallskip
\textit{\{\{base prompt\}\}} \par\smallskip
\texttt{\#\# Reflect on Tool Results} \\
After your previous commands executed, you MUST pause to reflect on the outcome before proceeding. Analyze the command's output and compare this against your expected result. Critically evaluate if the environment has changed as you anticipated. Determine if an additional command is necessary to explicitly verify the outcome of the previous command, similar to how a human would run \texttt{ls -l} after \texttt{chmod +x file.txt} to confirm the permission change. Based on this reflection and verification, decide whether the last command was successful, requires a corrective action, or if you can confidently move on to the next step to achieve the overall goal.

\texttt{\#\# Exploration} \\
Begin with environment exploration: Before you can solve the task, you must understand the environment that you are in. This includes understanding the file system and available tools. After environment exploration, you must investigate every file you discovered. Do not proceed with the main task until you have examined the available resources.

\medskip\noindent\rule{\linewidth}{0.2pt}\medskip

\textbf{base prompt + exploration + investigate all files} \par\smallskip
\textit{\{\{base prompt\}\}} \par\smallskip
\texttt{\#\# Exploration} \\
Begin with environment exploration: Before you can solve the task, you must understand the environment that you are in. This includes understanding the file system and available tools.

After environment exploration, you must investigate every file you discovered. Do not proceed with the main task until you have examined the available resources.
\end{tcolorbox}
\caption{Prompt variants evaluated on Terminal-Bench (Table~\ref{tab:appendix:tbench_all_prompts}). Each variant is a complete prompt; \textit{\{\{base prompt\}\}} denotes the shared base prompt that is identical across all conditions.}
\label{fig:appendix:prompt_variants}
\end{figure}

% ----------------------------------------------------------------------------------------------
\section{SFT Training Data Details}
\label{appendix:sft_training_data_details}

All fine-tuned models are trained from a variant of \texttt{command-a-reasoning}~\citep{command_a, cohere2025commandareasoningblog} that has been supervised fine-tuned for improved instruction following.

\paragraph{Rejection sampling.}
For each training task instance, we generate five agent trajectories using \texttt{gpt-oss-120b} with high reasoning. Each trajectory is a complete multi-turn interaction between the agent and the environment, consisting of alternating reasoning, action, and observation turns. We retain only the shortest successful trajectory per task. This yields training data that is both correct and concise, avoiding unnecessarily long solution paths. We use Terminus 1 as the agent to generate these trajectories.

\paragraph{Training details.}
We train three models on different task distributions:
\begin{itemize}
    \item \textbf{T-Bench-SFT}: Trained on Terminal-Bench-like tasks sourced from an external vendor, covering a broad distribution of terminal-based tasks. 14{,}005 trainable turns\footnote{A trainable turn is a single assistant message (comprising reasoning and action) within a trajectory.}, trained for 2 epochs.
    \item \textbf{AppWorld-SFT}: Trained on tasks from the AppWorld training split, covering API-based digital tasks. 15{,}841 trainable turns, trained for 2 epochs.
    \item \textbf{SWE-Bench-SFT}: Trained on tasks from SWE-smith, covering code editing and software engineering tasks. 21{,}424 trainable turns, trained for 1.5 epochs.
\end{itemize}
We chose the number of epochs such that all models are trained on approximately 30k effective task-specific turns (turns $\times$ epochs). To prevent overfitting to the task-specific data, we include a general-purpose tool-use SFT mixture as auxiliary data with a 1:1 mixing ratio in each training run. Training only on the general-purpose SFT mixture achieves $5.12$ pass@1 on Terminal-Bench and $0.18$ pass@1 on AppWorld. This near-zero baseline confirms that the agentic capability and environmental curiosity observed in our fine-tuned variants is attributable to the task-specific training data.

\section{AppWorld is a subset of Terminal-Bench}
\label{app:appworld-subset-of-tbench}
AppWorld tasks require an agent to discover, call, and reason over APIs across nine simulated day-to-day applications (e.g., Amazon, Spotify, Venmo) in a multi-turn fashion, spanning 457 endpoints in total. This task type is also present in terminal-bench but accounts for only a small slice of its distribution. Specifically, four of T-Bench v1's 80 tasks exercise the same core loop of API endpoint discovery, reasoning and interaction:
\begin{itemize}
    \item \texttt{simple-sheets-put} asks the agent to query a spreadsheet API to extract structured data, and then issue the correct sequence of API calls to create a spreadsheet, add a named sheet, populate cells with tabular data, and compute a derived column.
    \item \texttt{simple-web-scraper} asks the agent to scrape structured data from a web service, extract fields from HTML, aggregate the results into a CSV, and produce a summary report.
    \item \texttt{create-bucket} involves configuring an S3 bucket via CLI-based API calls
    \item \texttt{security-vulhub-minio} requires the agent to interact with a running MinIO service via its API to extract its credentials.
\end{itemize}

These four tasks, i.e. 5\% of T-Bench v1 tasks, share AppWorld's defining pattern but represent only a narrow slice of terminal-bench's broader distribution. Per-task complexity is comparable across the two benchmarks: on successful trajectories, \texttt{gpt-oss (120b)} with high reasoning averages 26.2 turns on terminal-bench and 29.6 on AppWorld. For reference, SWE-Bench requires 62.8 turns on average, reflecting substantially different per-task complexity. The combination of terminal-bench containing AppWorld-style tasks and both benchmarks exhibiting similar per-task complexity supports our view of AppWorld as a narrow subset of the T-Bench v1 task distribution.

% ----------------------------------------------------------------------------------------------
\section{Agent Implementations}
\label{appendix:agent_implementations}
We evaluate Terminus~1~\citep{terminal-bench} and SWE-agent~\citep{sweagent}. These agents differ in many aspects. The most important factors are prompts, how large outputs are truncated, and how commands are executed with SWE-agent using blocking commands while Terminus allows the model to specify timeouts and interrupt long-running processes. Among the many differences in implementation is also how tool results are formatted, i.e., how the terminal history is presented. We use the default Terminus and SWE-agent implementations but adapt them to use native function-calling APIs over raw prompting to remove the potential variable of out-of-distribution function calling interfaces (introduced in proprietary scaffolds) to instead rely on a provider's native tool-use interface.

\begin{figure}[h]
\noindent{\small\textbf{Terminal-Bench System Prompt}}
\vspace{4pt}
\begin{tcolorbox}[
  blanker, borderline west={1.5pt}{0pt}{gray!60},
  left=8pt, top=4pt, bottom=4pt
]
\centering

\begin{lstlisting}[style=prompt]
You are an AI assistant tasked with solving command-line tasks in a Linux environment. Your goal is to solve the task by iteratively interacting with a terminal session using the provided "terminal_use" tool.

As an example, its "commands" parameter can look like:
{"commands": [{"keystrokes": "ls -la\n", "wait": 0.1}, {"keystrokes": "cd project\n", "wait": 0.1}]}

## keystrokes
- Each "keystrokes" will be sent to the terminal completely verbatim via `tmux send-keys`, so write them exactly as you want it.
- Ctrl keys may be prefixed with 'C-' or '^', Shift keys with 'S-' and Alt (meta) with 'M-'. In addition, the following special key names are accepted: Up, Down, Left, Right, BSpace, BTab, DC (Delete), End, Enter, Escape, F1 to F12, Home, IC (Insert), NPage/PageDown/PgDn, PPage/PageUp/PgUp, Space, and Tab.
- If it's an actual command you want to execute (e.g. "ls"), make sure to end it with a newline (\n) which signals "enter".
- Do not include extra whitespaces before or after the keystrokes unless necessary.
- only include multiple commands at a time when you expect the commands to finish almost instantly (like cd), otherwise use one command at a time!

## wait
- "wait" specifies the number of seconds to wait before the next "keystrokes" will be sent or, in case it's already the last one, the seconds to wait before the terminal screen will get captured (via `tmux capture-pane`) and returned as tool result.
- For slow commands (e.g. make, python3 [long running script], wget [file]), allow enough time for execution.
- Do NOT wait longer than 60 seconds in one go. Prefer polling in shorter intervals to see intermediate status.

## "terminal_use" result
- After all the commands are sent and the wait is over, you'll see the latest terminal status in "terminal_status". It will show either "Current Terminal Screen" which is obtained by `tmux capture-pane -p` and contains only the visible contents of the pane, or "New Terminal History" which is obtained by `tmux capture-pane -p -S -` (i.e. capturing the full history) and then dropping the old history that has previously been seen.
- The "in_progress" attribute signals whether the terminal screen is still receiving new contents when capture-pane happens (e.g. when "wait" is over before the command completes execution). If you want to keep waiting to receive a fresher status, send {"keystrokes": ""} with some additional wait time.
- The "truncated" attribute signals whether "terminal_status" gets truncated in the middle. If it does, it will also mention how many bytes get truncated.

IMPORTANT
- You must complete the task all by yourself without ever asking the user for clarification or help.
- Make sure you use "terminal_use" to complete the task and actually get things done. You have root access to the terminal and can install packages, edit files and execute programs.
- The user wants you to GET THINGS DONE on their behalf. They do NOT want you to suggest solutions to them in the response but instead you must implement the soluion using commands. Make sure you actually execute commands (including writing and running code or scripts) to complete the task.
- Only when you are absolutely certain the task has been successfully completed will you write your final response. You will be graded based on what happens in the terminal session, NOT your final response. So be concise and only write "DONE" at the end.

## Exploration
Begin with environment exploration: Before you can solve the task, you must understand the environment that you are in. This includes understanding the file system and available tools.
After environment exploration, you must investigate every file you discovered. Do not proceed with the main task until you have examined the available resources.
\end{lstlisting}
\end{tcolorbox}
\caption{The Terminal-Bench system prompt provided to the Terminus agent during evaluation of the Terminal-Bench benchmark.}
\label{appendix:fig:terminal-bench-prompt}
\end{figure}

\begin{figure}[h]
\noindent{\small\textbf{AppWorld System Prompt}}
\vspace{4pt}
\begin{tcolorbox}[
  blanker, borderline west={1.5pt}{0pt}{gray!60},
  left=8pt, top=4pt, bottom=4pt
]
\centering
\begin{lstlisting}[style=prompt]
You are an AI assistant tasked with completing day-to-day tasks by writing code to interact with apps through their APIs.
You can interact with the API by calling the `cli` command. Your goal is to solve the task by iteratively interacting with a terminal session using the provided "terminal_use" tool.

As an example, its "commands" parameter can look like:
{"commands": [{"keystrokes": "cli --help\n", "wait": 2.0}, {"keystrokes": "ls -a\n", "wait": 2.0}]}

## keystrokes
- Each "keystrokes" will be sent to the terminal completely verbatim via `tmux send-keys`, so write them exactly as you want it.
- Ctrl keys may be prefixed with 'C-' or '^', Shift keys with 'S-' and Alt (meta) with 'M-'. In addition, the following special key names are accepted: Up, Down, Left, Right, BSpace, BTab, DC (Delete), End, Enter, Escape, F1 to F12, Home, IC (Insert), NPage/PageDown/PgDn, PPage/PageUp/PgUp, Space, and Tab.
- If it's an actual command you want to execute (e.g. "ls"), make sure to end it with a newline (\n) which signals "enter".
- Do not include extra whitespaces before or after the keystrokes unless necessary.
- only include multiple commands at a time when you expect the commands to finish almost instantly (like cd), otherwise use one command at a time!

## wait
- "wait" specifies the number of seconds to wait before the next "keystrokes" will be sent or, in case it's already the last one, the seconds to wait before the terminal screen will get captured (via `tmux capture-pane`) and returned as tool result.
- For slow commands (e.g. make, python3 [long running script], wget [file]), allow enough time for execution.
- Do NOT wait longer than 60 seconds in one go. Prefer polling in shorter intervals to see intermediate status.

## "terminal_use" result
- After all the commands are sent and the wait is over, you'll see the latest terminal status in "terminal_status". It will show either "Current Terminal Screen" which is obtained by `tmux capture-pane -p` and contains only the visible contents of the pane, or "New Terminal History" which is obtained by `tmux capture-pane -p -S -` (i.e. capturing the full history) and then dropping the old history that has previously been seen.
- The "in_progress" attribute signals whether the terminal screen is still receiving new contents when capture-pane happens (e.g. when "wait" is over before the command completes execution). If you want to keep waiting to receive a fresher status, send {"keystrokes": ""} with some additional wait time.
- The "truncated" attribute signals whether "terminal_status" gets truncated in the middle. If it does, it will also mention how many bytes get truncated.

IMPORTANT
- You must complete the task all by yourself without ever asking the user for clarification or help.
- Make sure you use "terminal_use" to complete the task and actually get things done. You have root access to the terminal and can install packages, edit files and execute programs.
- The user wants you to GET THINGS DONE on their behalf. They do NOT want you to suggest solutions to them in the response but instead you must implement the soluion using commands. Make sure you actually execute commands (including writing and running code or scripts) to complete the task.
- Only when you are absolutely certain the task has been successfully completed will you write your final response. You will be graded based on what happens in the terminal session, NOT your final response. So be concise and only write "DONE" at the end.

Before attempting to complete the task, you must discover what APIs are available to you by calling `cli --help`
\end{lstlisting}
\end{tcolorbox}
\caption{The AppWorld system prompt provided to the Terminus agent during evaluation of the AppWorld benchmark.}
\label{appendix:fig:appworld-bench-prompt}
\end{figure}

\begin{figure}[h]
\noindent{\small\textbf{SWE-Bench System Prompt}}
\vspace{4pt}
\begin{tcolorbox}[
  blanker, borderline west={1.5pt}{0pt}{gray!60},
  left=8pt, top=4pt, bottom=4pt
]
\centering
\begin{lstlisting}[style=prompt]
You are a helpful assistant that can interact with a computer to solve tasks.
I've uploaded a python code repository in the directory /testbed
Your goal is to solve the pull request by iteratively interacting with a terminal session using the provided "terminal_use" tool.

As an example, its "commands" parameter can look like:
{"commands": [{"keystrokes": "ls -la\n", "wait": 2.0}, {"keystrokes": "cd project\n", "wait": 2.0}]}

## keystrokes
- Each "keystrokes" will be sent to the terminal completely verbatim via `tmux send-keys`, so write them exactly as you want it.
- Ctrl keys may be prefixed with 'C-' or '^', Shift keys with 'S-' and Alt (meta) with 'M-'. In addition, the following special key names are accepted: Up, Down, Left, Right, BSpace, BTab, DC (Delete), End, Enter, Escape, F1 to F12, Home, IC (Insert), NPage/PageDown/PgDn, PPage/PageUp/PgUp, Space, and Tab.
- If it's an actual command you want to execute (e.g. "ls"), make sure to end it with a newline (\n) which signals "enter".
- Do not include extra whitespaces before or after the keystrokes unless necessary.
- only include multiple commands at a time when you expect the commands to finish almost instantly (like cd), otherwise use one command at a time!

## wait
- "wait" specifies the number of seconds to wait before the next "keystrokes" will be sent or, in case it's already the last one, the seconds to wait before the terminal screen will get captured (via `tmux capture-pane`) and returned as tool result.
- For slow commands (e.g. make, python3 [long running script], wget [file]), allow enough time for execution.
- Do NOT wait longer than 60 seconds in one go. Prefer polling in shorter intervals to see intermediate status.

## "terminal_use" result
- After all the commands are sent and the wait is over, you'll see the latest terminal status in "terminal_status". It will show either "Current Terminal Screen" which is obtained by `tmux capture-pane -p` and contains only the visible contents of the pane, or "New Terminal History" which is obtained by `tmux capture-pane -p -S -` (i.e. capturing the full history) and then dropping the old history that has previously been seen.
- The "in_progress" attribute signals whether the terminal screen is still receiving new contents when capture-pane happens (e.g. when "wait" is over before the command completes execution). If you want to keep waiting to receive a fresher status, send {"keystrokes": ""} with some additional wait time.
- The "truncated" attribute signals whether "terminal_status" gets truncated in the middle. If it does, it will also mention how many bytes get truncated.

## IMPORTANT
- You must complete the task all by yourself without ever asking the user for clarification or help.
- Make sure you use "terminal_use" to complete the task and actually get things done. You have root access to the terminal and can install packages, edit files and execute programs.
- The user wants you to GET THINGS DONE on their behalf. They do NOT want you to suggest solutions to them in the response but instead you must implement the soluion using commands. Make sure you actually execute commands (including writing and running code or scripts) to complete the task.
- Only when you are absolutely certain the task has been successfully completed will you write your final response. You will be graded based on what happens in the terminal session, NOT your final response. So be concise and only write "DONE" at the end.

## General Task Instructions
I've already taken care of all changes to any of the test files described in the PR description. This means you DON'T have to modify the testing logic or any of the tests in any way!
Your task is to make the minimal changes to non-tests files in the /testbed directory to ensure the PR description is satisfied.
Follow these steps to resolve the issue:
1. As a first step, it might be a good idea to find and read code relevant to the PR description
2. Create a script to reproduce the error and execute it with `python <filename.py>` using the terminal?use, to confirm the error
3. Edit the sourcecode of the repo to resolve the issue
4. Rerun your reproduce script and confirm that the error is fixed!
5. Think about edgecases and make sure your fix handles them as well
Your thinking should be thorough and so it's fine if it's very long.
\end{lstlisting}
\end{tcolorbox}
\caption{The SWE-Bench system prompt provided to the Terminus agent during evaluation of the SWE-Bench benchmark. This prompt closely follows the SWE-agent prompt from~\citet{sweagent}, adapted to our scaffold's \texttt{terminal\_use} interface.}
\label{appendix:fig:swe-bench-prompt}
\end{figure}

\end{document}